# NEURAL NETWORK BASED CLASSIFICATION OF BONE METASTASIS BY PRIMARY CARCINOMA

Marija Prokopijević[1], Aleksandar Stančić[2], Jelena Vasiljević[1,2],
Željko Stojković[1], Goran Dimić[1], Jelena Sopta[4], Dalibor Ristić[2] and
Dhinaharan Nagamalai[5]

[1]Institute Mihajlo Pupin, University, Belgrade, Serbia
[2]School of Computing, Union University, Belgrade, Serbia
[4]Medical Faculty, Institute of Pathology, University of Belgrade
[5]Wireilla, Australia

## ABSTRACT

*Neural networks have been known for a long time as a tool for different types of classification, but only just in the last decade they have showed their entire power. Along with appearing of hardware that is capable to support demanding matrix operations and parallel algorithms, the neural network, as a universal function approximation framework, turns out to be the most successful classification method widely used in all fields of science. On the other side, multifractal (MF) approach is an efficient way for quantitative description of complex structures [1] such as metastatic carcinoma, which recommends this method as an accurate tool for medical diagnostics. The only part that is missing is classification method. The goal of this research is to describe and apply a feed-forward neural network as an auxiliary diagnostic method for classification of multifractal parameters in order to determine primary cancer.*

## 1. INTRODUCTION

It has been shown that among microscopic medical images could be found statistically significant differences. This result promises to be powerful base point in the process of determining primary carcinoma in cases of intraoseal metastatic carcinoma. The ultimate goal of multifractal analysis is prompt determination of primary carcinoma and assistance in medical diagnostics by reduction of any subjective factor and minimizing of the error function in the process of the classification [1]. The main obstacle in the application of the results obtained in [1] is lack of automation in the diagnostic process, as well as the inability of new samples classification. However, in order to determine the tumor primary location for a particular image, it is necessary to retrieve multifractal parameters. FracLac [2], the image processing and the multifractal analysis program, is the well-known tool meant for obtaining the multifractal parameters from grey scale images. Having retrieved multifractal parameters, they should be classified usually by some machine learning algorithm, such as single decision tree, random forest, SVM or neural network. The aim of this study is automation of the last part of the diagnostics, which refers to data classification by the neural network implementation and its automatic startup in order to properly assign class to arbitrary sample.

In order to implement multilayer feed-forward neural network, we have used Java and Octave as universal programming language and software package that significantly facilitate the stages of the implementation such as backpropagation algorithm.

## 2. APPLICATION OF FRACTALS ON IMAGES

In the second half of the twentieth century, mathematician Benoit Mandelbrot determined the regularity in the nature of the objects, called them fractals and defined the theory of surface roughness [3]. The roughness theory refers to natural forms, such as mountains, coasts and river basins, the structure of plants, blood vessels, lungs, which cannot be described by Euclidean geometry.

Natural objects and phenomena do not exhibit strict fractal properties, even when they are self-similar, but can have statistical self-similarity. For example, the structure of the seashore, the appearance of a relief or cloud, the structure of some biological systems or signals, exhibit self-similar properties, but in different scales the shape is not exactly the same [1].

While fractal analysis describes forms characterized by strict mathematical properties, the multifractal analysis defines fractal properties in natural objects and phenomena. Their application in medical diagnostics is presented in [4]. These results enable and encourage the further research in the field.

### 2.1. Multifractal application in medical image analysis

The cancer cell is one of the natural forms that can be expressed by fractals. It is characterized by chaotic, poorly regulated cell growth [1], which is not a characteristic of healthy organisms. A healthy cell defines a form that helps in their functioning, while the appearance of cancer cells is usually abnormal. Cancer cells do not have a specific function and their abnormality is, among the others, expressed by size that is either lower or higher than a healthy cell. Irregular growth also occurs in the nucleus and cytoplasm of the cell. That is, the nucleus of the malignant cells is bigger than healthy cells nucleus, while the cytoplasm is scarce and intensively colored or very pale [5].

Since the malignant cells show certain properties of self-similarity, they are suitable for the application of multifractal analysis.

The multifractal analysis of digital medical images obtained from [1] has been carried out using ImageJ [6] (image analysis) along with FracLac, (multifractal analysis) software package.

## 3. INTRAOSEAL METASTASES

The spread of cancer in the body emerges when cells have detached from the primary carcinogenic tumour and travel through the lymphatic system or bloodstream to other organs and causing the onset of metastases.

Metastases could appear in different parts of the body. Certain types of cancer, such as breast, prostate, lung, thyroid and kidney cancer, are most commonly spread on the bones [7].

More than two-thirds of patients with breast cancer, and about one-third with some other types of cancers, such as lung and kidney cancer, are exposed to risk of bone metastases. According to statistics, the mortality of this cancer is high. There is statistics that 69% of the patients who died of breast cancer had metastases on the bones [7].

The data subject to classification are obtained from the samples of the cancer metastases on the bones. Microscopic images of metastasis, depending on primary carcinoma, have their shape and statistically significant differences according to which further analysis and classification are carried out.

Digital images of three groups of metastases of intraoseal cancer were observed:

- metastatic carcinoma of renal cells, shown in Fig. 1,
- metastatic breast carcinoma, shown in Fig. 2,
- metastatic lung carcinoma, shown in Fig. 3.

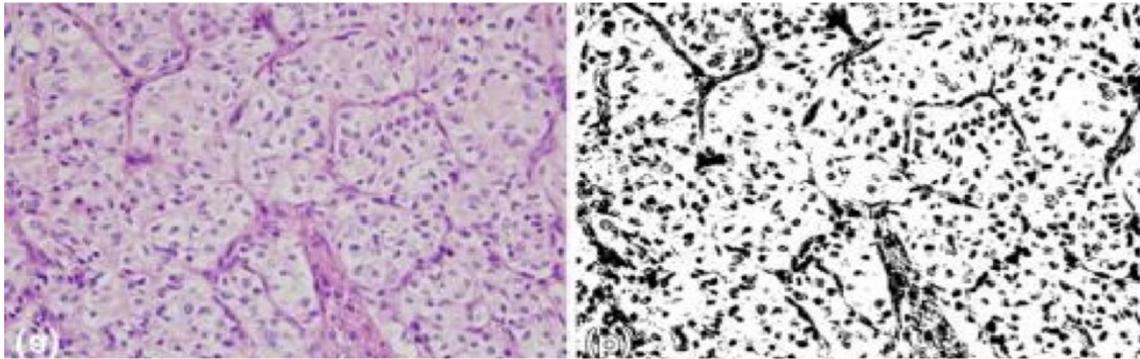

Figure 1. Metastatic renal cell carcinoma: (a) Microscopic image; (b) Binary image obtained using FracLac program

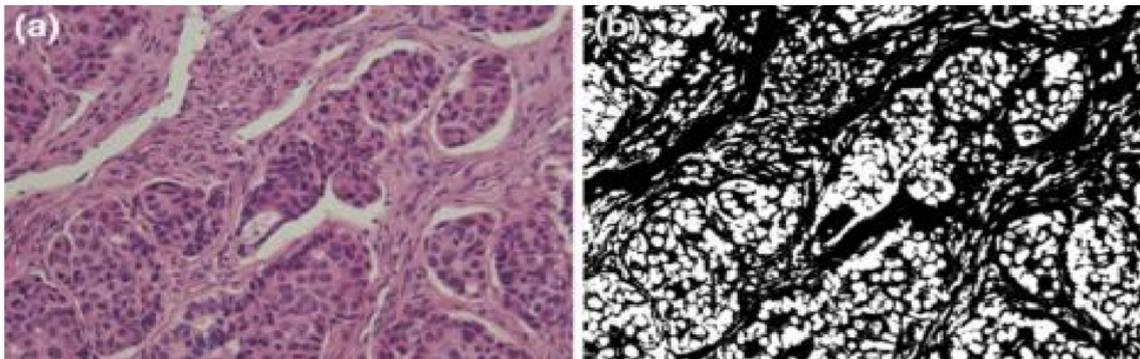

Figure 2. Metastatic breast carcinoma: (a) Microscopic image; (b) Binary image obtained using FracLac program

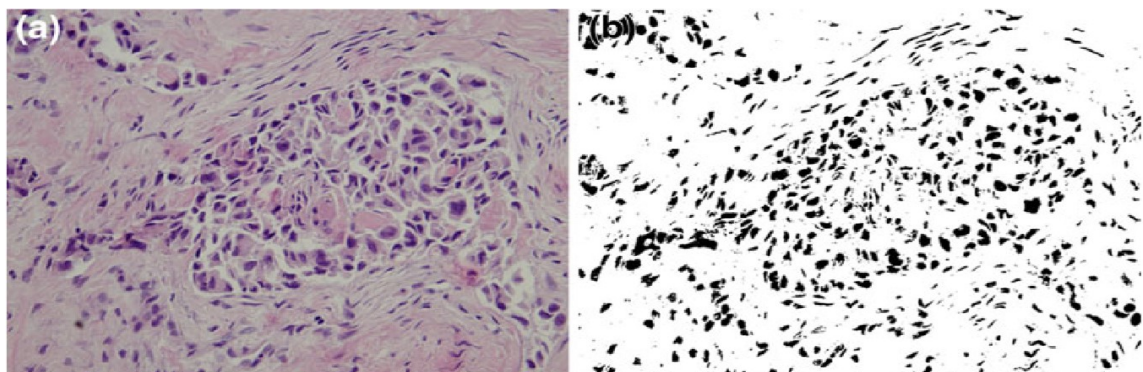

Figure 3. Metastatic lung carcinoma: (a) Microscopic image; (b) Binary image obtained using FracLac program

## 4. NEURAL NETWORKS AND CLASSIFICATIONS

The term "neural networks" first appeared in the 1940s, when scientists McCulloch and Pitts defined a mathematical model for the search of biological information [8]. Today, neural networks are used to solve many non-linear problems, such as pattern recognition, classification, clustering, regression, compression, robotics…

In order to successfully classify images, based on multifractal parameters, it is necessary to implement a multilayer neural network. Although a single-layer network is simpler to train, its precision is not great because it is able to classify only linearly separable data. According to Fig. 4 it is clear that the results of multifractal analysis are not linearly separable.

On the other hand, although the input data are not linearly separable, it is possible to transform them applying a nonlinear function in order to "become separable". Thus, the multilayer perceptron can solve a linearly inseparable problem, provided that it has an appropriate set of parameters (weights) and "nonlinearity" as an activation function. In other words, if a hidden layer of a neural network does not solve the problem, it is necessary to re-calibrate weights, as well as the number of neurons in the hidden layer of the neural network [9].

Generally, it is shown that the network with two layers of weight, i.e. three layers of neurons, is capable of approximating any continuous function. The only constraint is that the network diagram has to be feed-forward, so it does not contain feedback loops. In this way, it is ensured that network outputs are calculated as a function of inputs and weights [9].

In order to carry out this research, it has been implemented a multilayer neural network which contains an input, a hidden and an output layer of a neurons. The size of the input layer depends on the number of multifractal parameters, the size of output layer is determined by the number of classes (i.e. three), while the size of the hidden layer is adjusted to as accurately as possible approximate the desired function.

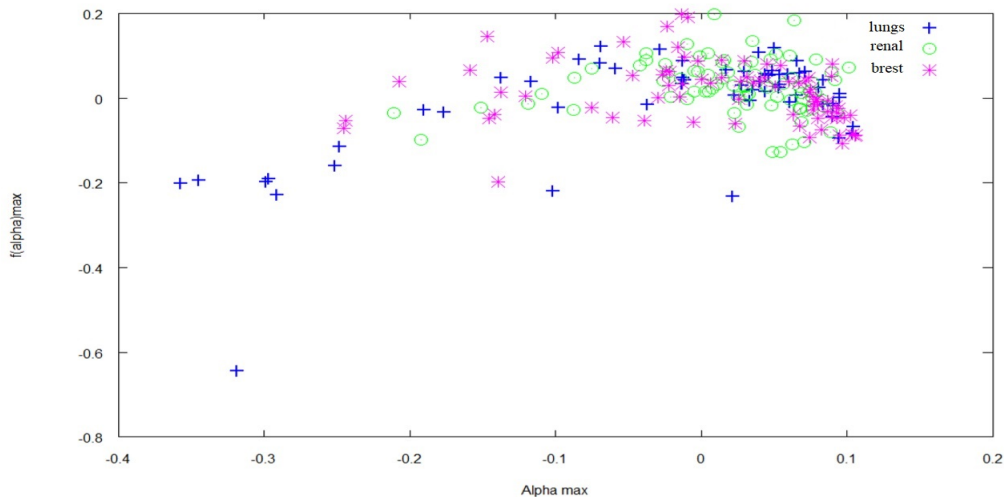

Figure 4. The graph shows the linear inseparability of the data

## 5. APPLICATION OF NEURAL NETWORKS IN MULTIFRACTAL ANALYSIS

### 5.1. Neural network input parameters

Having applied FracLac software, the following multifractal parameters are obtained:

- $D_{max}$ - maximum of generalized fractal dimension
- Q - exponent of the fractal dimension
- $α_{min}$ - minimum value of the Hõlder exponent
- $f(α)_{min}$ - minimum of multiftactal spectrum $f(α)$
- $α_{max}$ - maximum value of the Hõlder exponent
- $f(α)_{max}$ - maximum of multiftactal spectrum $f(α)$

According to the mean values of the parameters (Table 1), as well as the detailed analysis of the dependencies among them, it is concluded that the data are not linearly separable, as shown in Fig. 4.

**5.2. Neural network implementation**

A free programming language for numerical analysis, Octave which is compatible with MATLAB, and Java have been used to implement a feed-forward neural network and learning algorithm. The advantage of the Octave programming language is the simple execution of matrix operations, as well as built-in libraries for calculating the gradient.

Table 1. Average parameter values obtained using FracLac program [1].

| Variable | Group | Average |
|---|---|---|
| $D_{max}$ | Lung | 2.798677 |
| | Breast | 2.797801 |
| | Renal cells | 2.786872 |
| Q | Lung | -6.37214 |
| | Breast | -6.38071 |
| | Renal cells | -6.41573 |
| $α_{min}$ | Lung | 3.093834 |
| | Breast | 3.121177 |
| | Renal cells | 3.103109 |
| $f(α)_{min}$ | Lung | 0.758865 |
| | Breast | 0.740055 |
| | Renal cells | 0.762394 |
| $α_{max}$ | Lung | 1.975734 |
| | Breast | 1.993426 |
| | Renal cells | 1.993601 |
| $f(α)_{max}$ | Lung | 1.872967 |
| | Breast | 1.885918 |
| | Renal cells | 1.88775 |

The network is implemented as a feed-forward neural network with a backpropagation algorithm for weight learning. The accuracy of the network is measured by the cost function $J(\theta)$, represented by the equation (7), which is calculated during backpropagation after the feed-forward iteration.

The network input layer, denoted by $X = [x_i^{(j)}]$, represents the matrix of multifunctional parameters, where the indices $i$ and $j$ are related to the $i$-th parameter of the $j$-th sample. The dimension of the matrix is 27xm, where m is the number of samples in the training set. The input layer has 27 neurons, related to each multifractal parameter and all linear combinations of the parameters.

The network output layer, denoted by $Y = [y_k^{(j)}]$, represents the matrix of the output parameters. Each row $j$ of the matrix represents the output of single sample, where the $k$-th column represents the class $k$. Value $y_k^{(j)}$ will be 0 if the sample is not classified as carcinoma of class $k$, otherwise 1. All the classes are labelled with numbers 1, 2, and 3, where 1 represents breast cancer, 2 lung cancer, and 3 kidney cancer.

Parameters, i.e. the weights, that the neural network learns, are represented by the vectors $\theta^{(1)}$ and $\theta^{(2)}$, which are related to hidden and output layer respectively. There is a regularization parameter $\lambda$, which determines how much $\theta$ weights affect each step of learning. Also, $h_\theta(x)$ represents sigmoid function, while $s_1$, $s_2$ and $s_3$ are the size of the input, hidden and output layers.

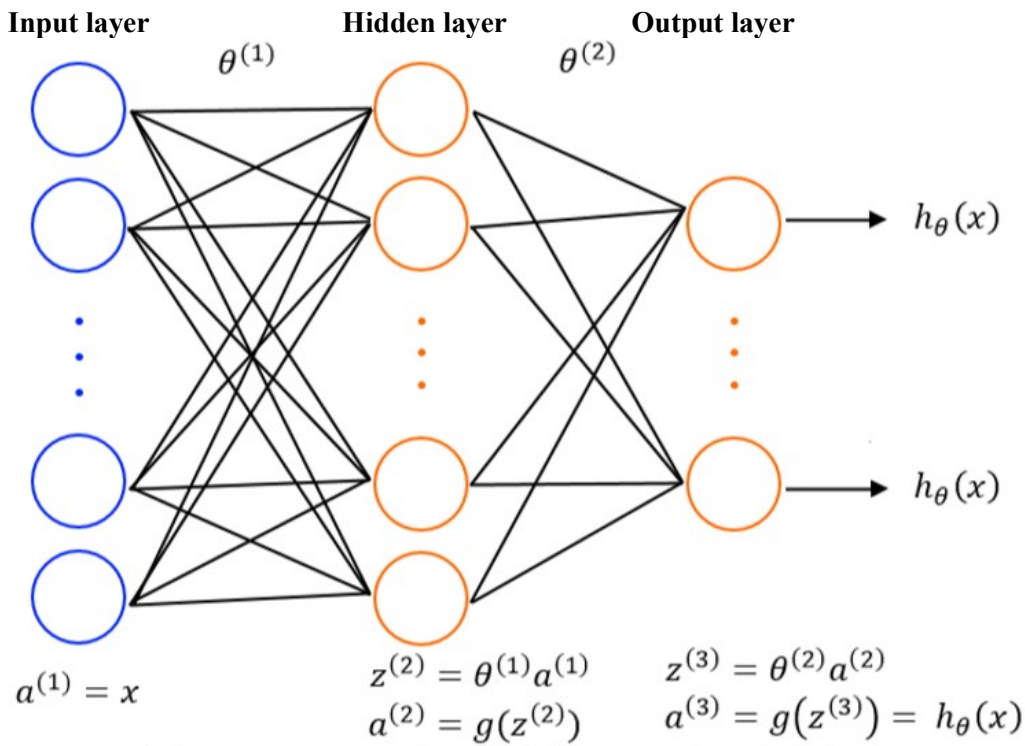

Figure 5. Feed-forward algorithm

The first part of the network learning refers to the feed-forward algorithm, which predicts the class for the given input based on the current weights. In this stage, the algorithm propagates values from the input to the output layer. The feed-forward algorithm is shown in Fig. 5.

The neural network uses the sigmoid function $h_\theta(x)$, given by the equation (1), to calculate the values that neurons send through the network. The sigmoid function value, shown in Fig. 6, is in the interval (0, 1) and represents an impulse which is sent to the next layer of the network.

$$h_\theta(x) = \frac{1}{1+e^{-\theta^T x}} \qquad (1)$$

Each neuron represents a logistic unit that calculates the value of the sigmoid function and sends the impulse to activate the next layer of neurons.

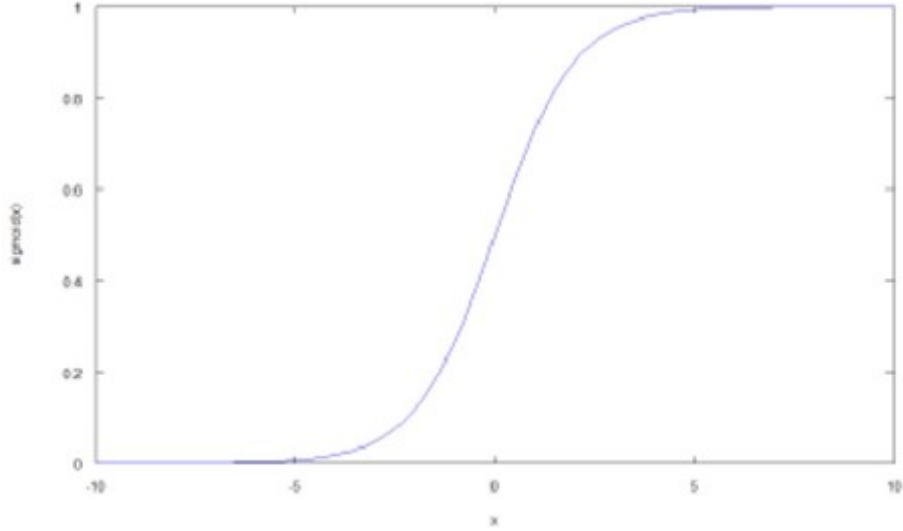

Figure 6. Sigmoid function

In order to calculate the impulse, each neuron calculates the activation function $a_i^{(j)}$, where $i$ represents an $i$-th activation unit in the $j$ layer. The activation function is represented by the sigmoid function $g(z)$, where $z$ is the linear combination of the input values $X$ and the weights $\theta$.

$$z = (\theta)^T * x \qquad (2)$$

$$g(z) = \frac{1}{1+e^{-z}} \qquad (3)$$

$$a_i^{(j)} = g\left(\theta_{i0}^{(j-1)} x_0 + \theta_{i1}^{(j-1)} x_1 + \ldots + \theta_{in}^{(j-1)} x_n\right) \qquad (4)$$

The previously mentioned cost function $J(\theta)$ is a measure of network precision. The next sentence describes the intuition originates from logistic regression. That is, the calculated cost value represents the penalty that the function will pay if the class is not correctly predicted. The function for individual cost calculation is defined as follows:

$$Cost(h_\theta(x), y) = \begin{cases} -\log(h_\theta(x)), & y = 1 \\ -\log(1 - h_\theta(x)), & y = 0 \end{cases} \qquad (5)$$

The cost is chosen to be a logarithmic function, as shown in Fig. 7 and 8, because the logarithm gives a convex function (the sigmoid function is not convex) and reflects the cost reduction intuition. Namely, if for a given sample $y = 1$, the sigmoid function tends to 1, the value of the cost function will tend to 0, i.e. the closer predicted and real value are the cost is lower.

Conversely, if y = 1 and the sigmoid value is 0, the cost function tends to infinity, i.e. the neural network will be highly punished because of wrong result.

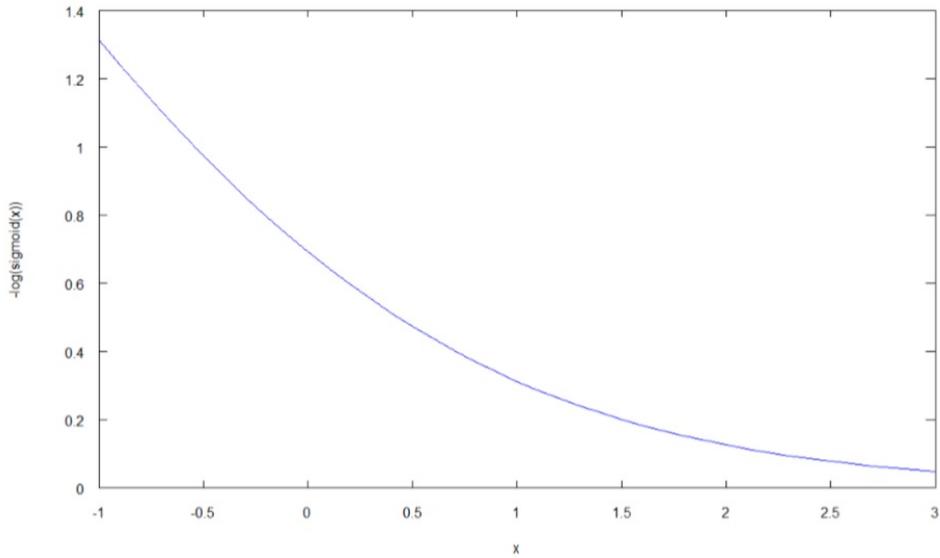

Figure 7. Cost function for y = 1

Similarly, if y = 0 and the value of the sigmoid function is 0, the cost function will tend to zero, and when the sigmoid tends to 1, the cost function tends to infinity.

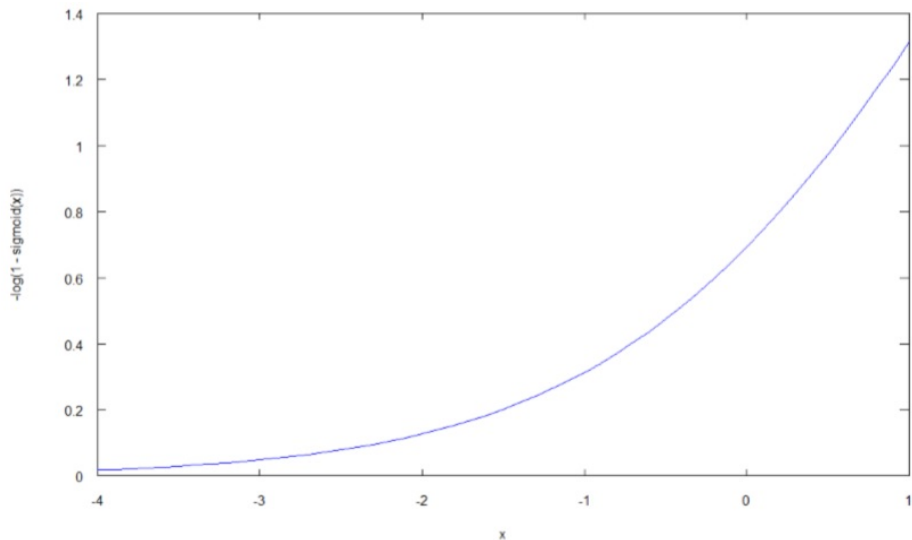

Figure 8. Cost function for y = 0

The cost function can be concisely represented by the following equation:

$$-y\log(h_\theta(x)) - (1-y)\log(1-h_\theta(x)) \quad (6)$$

So far, the cost function has been calculated only for a single sample. In order to calculate the total cost of the neural network, all the samples should be taken in consideration, as shown by the equation (7).

The first part of the equation (7) refers to the calculation of the cost function for all samples, where inner "Σ loop" takes in consideration all the classes for the individual sample and calculates costs for one by one sample, while the outer one sums sample costs. The second part of the equation refers to the regularization, and in order to avoid the network over-fitting, tends to keep as small as possible weights.

$$J(\theta) = \frac{1}{m}\sum_{i=1}^{m}\sum_{k=1}^{nk}\left[-y_k^{(i)}\log\left(\left(h_\theta(x^{(i)})\right)_k\right)-(1-y_k^{(i)})\log\left(1-\left(h_\theta(x^{(i)})\right)_k\right)\right]$$
$$+\frac{\lambda}{2m}\left[\sum_{j=1}^{S_2}\sum_{k=1}^{S_1}\left(\theta_{j,k}^{(1)}\right)^2+\sum_{j=1}^{S_3}\sum_{k=1}^{S_2}(\theta_{j,k}^{(2)})^2\right] \qquad (7)$$

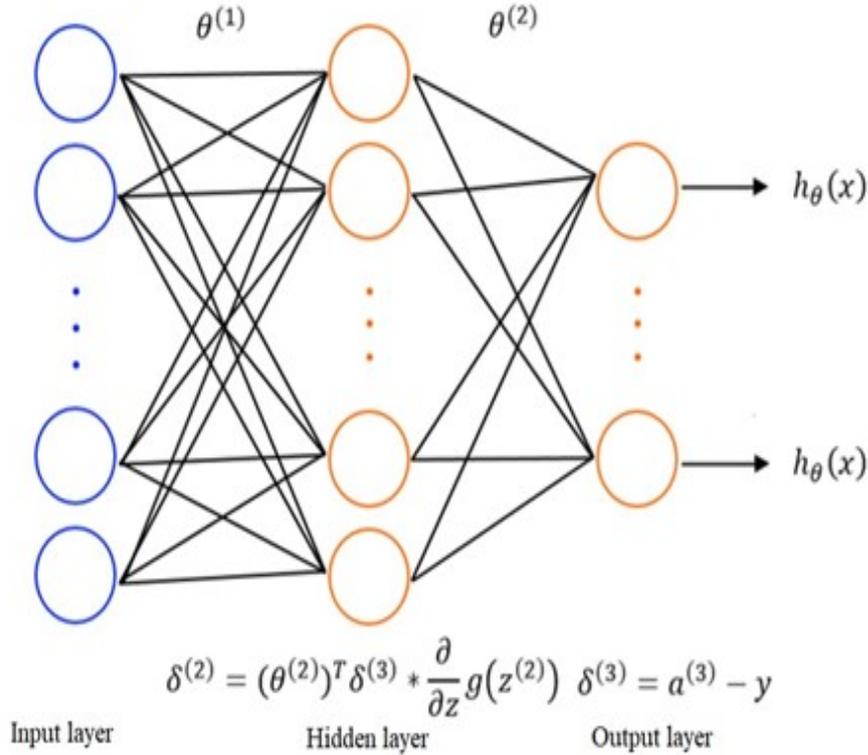

Figure 9. Backpropagation algorithm

As opposed to the feed-forward algorithm, backpropagation is calculated from the output layer to the input. The error is denoted as $\delta_i^{(j)}$, where $i$ represents the $i$-th unit of the layer $j$. Unlike the activation function, which is calculated in the same way for all layers, error calculation is dependent on the layers. Fig. 9 describes the backpropagation algorithm.

It is straightforward to calculate the error of the output layer. The error is obtained as the difference between the output values of the sample and the final results of the feed-forward algorithm:

$$\delta^{(3)} = a^{(3)} - y \qquad (8)$$

For the next layer of the network, it should be calculated derivatives of the function (i.e. found the function minimum) of the inner layer of the network.

$$\delta^{(2)} = (\theta^{(2)})^T \delta^{(3)} * \frac{\partial}{\partial z}g(z^{(2)}) \qquad (9)$$

The last step in the algorithm is updating of the weights. The feed-forward algorithm, followed by backpropagation will be repeated, until the cost function converges to a minimum or the predetermined number of iterations exceeds.

### 5.3. Parameters normalization and scaling

It is desirable to scale input data before training a neural network. Scaling is used when the input parameters have different ranges of values. For multifractal analysis, all six basic parameters are in different range of values, which causes an unfair parameter influences in the network itself. A parameter with a higher value has more impact on the network output, which might not describe real state.

One way of scaling is to transform values to the *N(0, 1)* distribution. During scaling, the data are regarded as independent of each other, and for each variable $x_i$, it is necessary to be calculated mean value $\mu_i$ as well as the variance $\sigma_i^2$. The final result is a normalized value $\overline{x_i}$.

$$\mu_i = \frac{1}{m} \sum_{n=1}^{m} x_i^{(n)} \qquad (10)$$

$$\sigma_i^2 = \frac{1}{m-1} \sum_{n=1}^{m} (x_i^{(n)} - \mu_i)^2 \qquad (11)$$

$$\overline{x_i^{(n)}} = \frac{x_i^{(n)} - \mu_i}{\sigma_i} \qquad (12)$$

## 6. RESULTS

The network has been trained on 1050 samples, 350 in each target group. Samples were provided from the Institute of Pathology, Medical Faculty, University of Belgrade. They represent original microscopic images analyzed by FracLac software in order to obtain the values of multifractal parameters, later used for classification purpose.

Before the network training, the data are divided into two sets, training and validation. The size ratio of these two sets is 75-25%, where 75% of the data are used for training the network, and 25% for the validation of the parameters.

### 6.1. Parameters for results estimation

The parameters used to estimate the results of the neural network are given in Table 2. They are described in details by the equations (13) - (18). The values are given for each target group (primary carcinoma) k separately. "True positive" value represents samples that are properly classified in their group k, "false positive" value corresponds to samples from group k, which are incorrectly classified into one of the other two groups. "False negative" value is related to samples that are not properly labelled and "true negative" value represents the samples correctly classified not to be part of group k.

Table 2. Parameters.

| Classification outcome | | Condition | | |
|---|---|---|---|---|
| | | Positive condition | Negative condition | |
| | Positive outcome | true positive (tp) | false positive (fp) | Precision |
| | Negative outcome | false negative (fn) | true negative (tn) | Negative predictive value |
| | | Sensitivity | Specificity | Accuracy |

$$Accuracy = \frac{tp + tn}{tp + tn + fp + fn} \quad (13)$$

$$Sensitivity = \frac{tp}{tp + fn} \quad (14)$$

$$Specificity = \frac{tn}{tn + fp} \quad (15)$$

$$Geometric\ mean\ of\ sensitivity\ and\ specificity = \sqrt{sens. * spec.} \quad (16)$$

$$Precision = \frac{tp}{tp + fp} \quad (17)$$

$$F - measure = \frac{2 * tp}{2 * tp + fp + fn} \quad (18)$$

## 6.2. Results

The precision of the classification obtained by the feed-forward neural network, with backpropagation, is given in Tables 3-5:

Table 3. Lungs results.

| Lung cancer | |
|---|---|
| Accuracy | 70.18% |
| Sensitivity | 54.55% |
| Specificity | 76.97% |
| Geometric mean sensitivity and specificity | 64.80% |
| Accuracy | 50.71% |
| F-Measure | 0.5255 |

Table 4. Renal results.

| Renal cancer | |
|---|---|
| Accuracy | 68.35% |
| Sensitivity | 52.48% |
| Specificity | 82.05% |
| Geometric mean sensitivity and specificity | 65.62% |
| Accuracy | 71.62% |
| F-Measure | 0.6057 |

Table 5. Brest results.

| Brest cancer | |
|---|---|
| Accuracy | 66.97% |
| Sensitivity | 50.98% |
| Specificity | 71.86% |
| Geometric mean sensitivity and specificity | 60.52% |
| Accuracy | 35.62% |
| F-Measure | 0.4194 |

## 7. CONCLUSIONS

Based on the obtained results, feed-forward neural network appears to be acceptable and applicable as an auxiliary diagnostic method for cancer classification, but it is definitely worth to invest an effort in trying to improve results and precision of the network even more. It is expected that deeper networks would give better results, so the research should be continued into the direction of applying deep learning networks. A properly structured recurrent neural network (RNN) might improve accuracy significantly. Since 2012, a convolutional neural network (CNN) is considered to be the ultimate method in image processing and classification, so CNN could be found an application in MF classification as well, and also it would be very interesting to compare results of the other methods with accuracy given by CNN.

### ACKNOWLEDGEMENTS

This work was supported by the Ministry of Education and Science, Republic of Serbia, Science and Technological Development grant III 43002.